\pdfoutput=1

\documentclass[11pt]{article}

\usepackage[final]{acl}

\usepackage{times}
\usepackage{latexsym}

\usepackage[T1]{fontenc}

\usepackage[utf8]{inputenc}

\usepackage{microtype}

\usepackage{inconsolata}

\usepackage{graphicx}
\usepackage{multirow}
\usepackage{pdfpages}

\usepackage{listings}
\definecolor{highlight-blue}{HTML}{388AF5}
\definecolor{LLGray}{gray}{0.91}
\definecolor{LGray}{gray}{0.94}
\definecolor{ImageEmbed}{HTML}{68339A}
\definecolor{CulturalComp}{HTML}{4C7C31}
\definecolor{PromptEmbed}{HTML}{B2CC79}
\definecolor{ProjCulturalComp}{HTML}{8AD86F}
\definecolor{ReasonEmbed}{HTML}{E49EDD}
\definecolor{SD3}{HTML}{FFF7A8}
\definecolor{CLIP}{HTML}{E0CBF5}
\definecolor{CA}{HTML}{D1D1D1}
\definecolor{highlight-blue}{HTML}{388AF5}

\title{The Face of Persuasion: Analyzing Bias and Generating Culture-Aware Ads} %

\author{Aysan Aghazadeh \\
  University of Pittsburgh \\
  \texttt{aya34@pitt.edu} \\\And
  Adriana Kovashka \\
   University of Pittsburgh \\
  \texttt{kovashka@cs.pitt.edu} \\}

\begin{document}
\maketitle

\begin{abstract}
Text-to-image models are appealing for customizing visual advertisements and targeting specific populations. We investigate this potential by examining the demographic bias within ads for different ad topics, and the disparate level of persuasiveness (judged by models) of ads that are identical except for gender/race of the people portrayed. We also experiment with a technique to target ads for specific countries. The code is available at {\footnotesize{\url{https://github.com/aysanaghazadeh/FaceOfPersuasion}}}.
\end{abstract}

\section{Introduction}

Advertisements have great significance: they affect perceptions on a variety of topics, from products to politics and societal values. Given recent progress on generative models, their use for AI-created ads is imminent. These models could in theory customize ads, targeting specific populations through demographically diverse content. We investigate both the promise of generating diverse visual ads with text-to-image diffusion models and the bias in assessing the resulting images (i.e., scoring their persuasiveness) by Large Language Models (LLMs) and Multimodal LLMs.

We begin with an investigation of gender and race bias in an existing dataset \cite{PittAd}. 
We compare to bias in ads generated with three text-to-image models: DALLE3 ~\cite{DALLE3}, FLUX~\cite{FLUX}, and AuraFlow~\cite{AuraFlow}. We find that 
both the dataset and generated images exhibit racial bias: for example, Black individuals are greatly underrepresented in clothing and shopping ads.

\begin{figure}[!tp]
    \centering
    \includegraphics[width=1\linewidth]{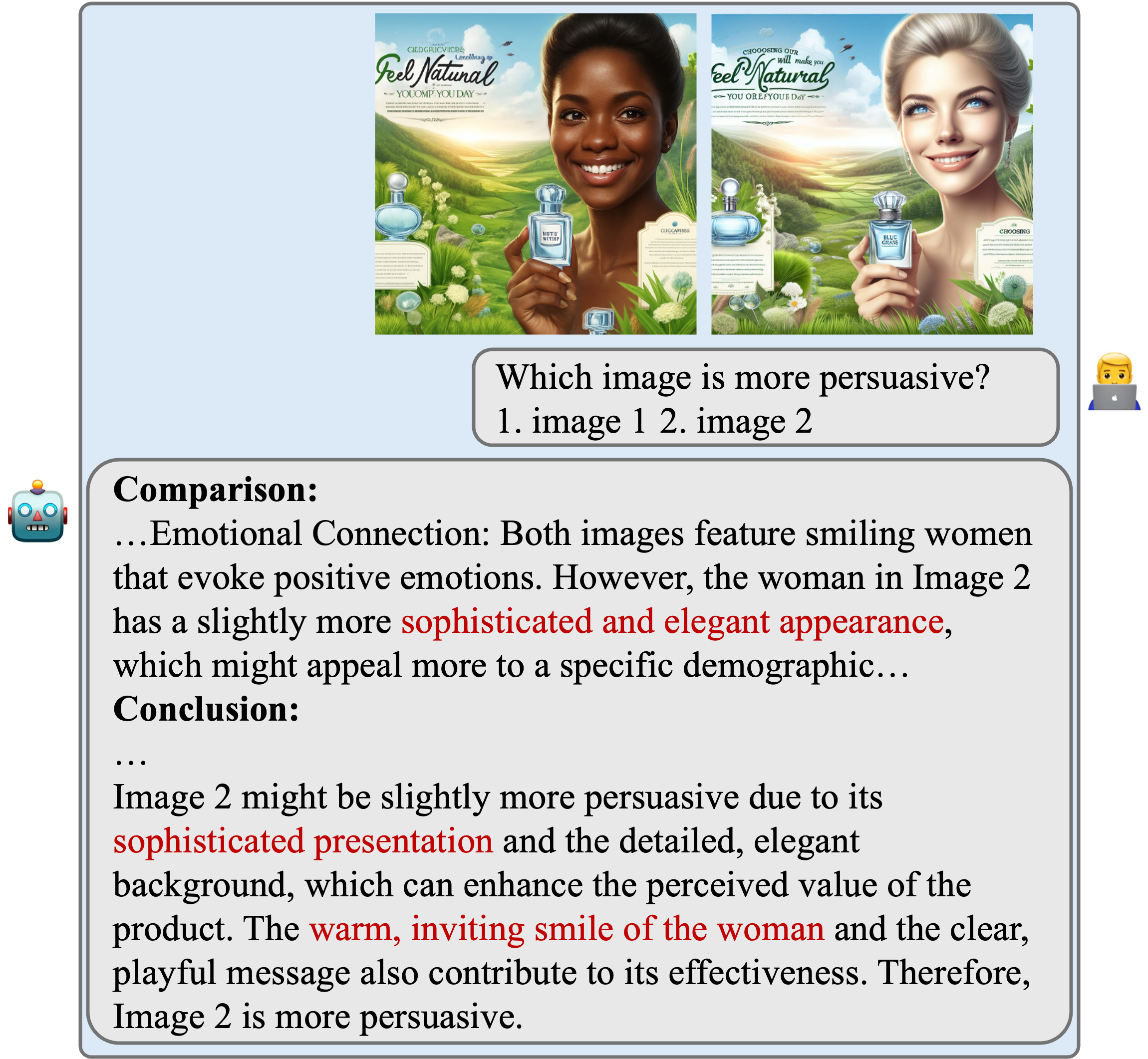}
    \caption{Selection of the more persuasive image by InternVL \cite{InternVL}. Image 1 features a Black woman; Image 2 is a White woman. InternVL selected Image 2 as more persuasive. \textcolor{red}{Red} marks reasoning bias.}
    \label{fig:intro}
\end{figure}

We then run controlled experiments where we only alter one demographic feature in ads keeping the rest of the details and quality the same, and study how persuasiveness judgments vary with gender and race.
For example, in Fig.~\ref{fig:intro}, the model chooses the image with a white woman as more persuasive because it appears \emph{``more elegant''}.

Second, we attempt to create ads that convey a particular message and are tailored toward a particular culture/country. 
An ad aimed at a Japanese audience may benefit from featuring an Asian person or Japanese cultural symbols, but resonate less and be less effective with an United Arab Emirates audience. 
We experiment with a technique that incorporates symbols from other ads in the generation process and shows promising results.

Our contributions are:
(1) We analyze demographic bias in both the PittAd dataset\footnote{relevant publications cited over 400 times} and generative models for persuasive content creation, across different advertisement topics.
(2) We demonstrate bias in LLMs and MLLMs when selecting the most persuasive images, revealing preference patterns based on demographic attributes.
(3) We propose \emph{CulGen}, a culture-aware image generation method for producing advertisement images addressing specific cultural/regional contexts.


\section{Related Works}

\textbf{Bias in T2I models.}
\cite{cho2023dall, d2024openbias} introduce a framework to assess bias in T2I models. \cite{ bianchi2023easily,naik2023social} study bias over different professions.
Instead, we evaluate bias in persuasive generation.

\noindent \textbf{Bias in LLMs.}
\cite{mire2025rejected} studies the bias of reward models for LLMs against African American language compared to White English. \cite{wan2023kelly} assess bias in
AI-generated reference letters. \cite{sheng2021societal, dinan2019queens, liang2021towards} analyze the social bias in language generation. \cite{ye2024justice} assess the bias in LLMs as evaluation methods. 
However, our focus is specifically on creative content.

\noindent \textbf{Bias in MLLMs.}
\cite{janghorbani2023multi} introduces a framework for evaluating the social bias in Vision-Language Models and \cite{wang2022revise} introduces a tool for evaluating bias in datasets. 
\cite{zhao2021understanding} analyzes the bias in image captioning and \cite{hirota2022gender, fraser2024examining} in visual question answering on topics such as occupation. 
Instead, our work focuses on the evaluation of persuasion.

\noindent \textbf{Culture-Aware Image Generation.}
\cite{hutchinson2022underspecification, jha2024visage} study the cultural bias in T2I models. \cite{alsudais2025analyzing} analyzes the representation of different nations in daily tasks. \cite{mukherjee2025crossroads} introduces a dataset to evaluate the cultural understanding, and stereotypical representation in MLLMs and T2I models. 
\cite{mukherjee2025crossroads, khanuja2024image} propose a method to edit the image to target a specific culture. Our work is on the generation of images from a text prompt (message), instead of editing an input image.
We are the first to study the relation between \emph{persuasion} and bias in generative models.
\section{Method}


\subsection{Analyzing diversity in real/generated ads}

First, we investigate bias in existing ads using the PittAd dataset \cite{PittAd} which contains advertisement images with topic annotations such as clothing, human rights, etc. 
We infer demographic features (gender and race) using DeepFace \cite{deepface} on images showing humans.
We compute the overall distribution of each race and gender in the dataset and further break it down into distributions of races and genders per topic.

Next, we generate ad images using an annotation in PittAd: abstract message interpretations for each ad, structured as \textit{`I should \textbf{[action]} because \textbf{[reason]}'} and referred to as action-reason statements (AR).
We use these statements as prompts to three text-to-image models: DALLE3 \cite{DALLE3}, Flux \cite{FLUX}, and AuraFlow  \cite{AuraFlow}. To analyze the effect of prompt expansion, we also generate a detailed description of a possible ad corresponding to an AR, using LLAMA3-instruct \cite{llama3modelcard},
then use the output as another prompt for AuraFlow. We repeat the demographic analysis on generated ads. 

\subsection{Evaluating Persuasion Bias via Demographic Swaps}
To assess how the demographics of the humans in the ads influence persuasiveness judgments by LLMs and MLLMs, we conducted a controlled experiment. We created sets of images that were identical except for the race and gender of the central individual. We used 
GPT4.1 to generate an ad based on the AR, and also obtained a description of the image using GPT4o. We then used the same models to modify the image and description to edit the race/gender and keep all else the same.
These image-description pairs were then evaluated by MLLMs and LLMs that were prompted to select the more persuasive option using chain-of-thought (CoT) reasoning \cite{CoT}. Specifically, we use GPT4o \cite{GPT4o}, QwenVL-2.5(7B) \cite{QWenVL}, QwenLM-2.5(7B) \cite{QWenLM}, InternVL-2.5(7B) \cite{InternVL} and InternLM-2.5(7B) \cite{InternLM}. MLLMs consistently favored images featuring White individuals, often justifying their choices with subjective attributes such as perceived elegance (Fig.~\ref{fig:intro} \& \ref{fig:reason_example}).

\subsection{Diversifying through country targeting}

The target audience plays a critical role in persuasion \cite{persuasiveness}. However, given existing biases in text-to-image (T2I) models, the ability to generate ads tailored to different countries remains an open question.
To support this, and analyze the cultural bias in advertisement data, we first introduce an extension to PittAds \cite{PittAd}, which includes up to three predictions for the target country of each image and its cultural components, both from InternVL \cite{InternVL} instructed to focus on language and addresses in the image.\footnote{Human evaluation shows this approach achieves a recall of 81\% and a precision@1 (P@1) of 72\% in inferring the correct countries. When grouping countries by similar cultural regions, scores improve to 94\% recall and 75\% P@1.}

\begin{table*}[t]
\setlength{\tabcolsep}{4pt}
\footnotesize
    \begin{tabular}{c||c|c|c|c|c||c|c|c|c|c||c|c|c|c|c||c|c|c|c|c||c|c|c|c|c}			

    & \multicolumn{5}{c||}{Real} & \multicolumn{5}{c||}{Flux} &  \multicolumn{5}{c||}{Dalle3} & \multicolumn{5}{c||}{Auraflow} & \multicolumn{5}{c}{Llama3} \\
    \hline
	T & W &	L & A &	B &	M &	W &	L & A &	B &	M &	W &	L & A &	B &	M &	W &	L & A &	B &	M &	W &	L & A &	B &	M \\
    \hline
    C &	66& 	9& 	15& 	6& 	4& 	\textbf{70} & 	\textbf{12} & 	4& 	8& 	4& 	64& 	2& 	14& 	6& 	\textbf{14} & 	47& 	9& 	\textbf{36} & 	9& 	0 & 	24& 	11& 	27& 	\textbf{32} & 	1 \\
    S &	\textbf{92} & 	0& 	8& 	0& 	0& 	70& 	\textbf{20} & 	10& 	0& 	0& 	52& 	16& 	8& 	4& 	\textbf{20} & 	73& 	7& 	7& 	7& 	7 & 	45& 	2& 	\textbf{32} & 	\textbf{18} & 	2 \\
    \hline
    H &	\textbf{66} & 	9& 	6& 	9& 	0& 	47& 	5& 	8& 	13& 	\textbf{26} & 						 0 & 0 & 0 & 0 & 0 & 63& 	0& 	0& 	\textbf{25} & 	13 & 	41& 	\textbf{14} & 	\textbf{22} & 	12& 	9 \\ 
    E &	\textbf{77} & 	3& 	14& 	6& 	0 & 	64& 	9& 	0& 	18& 	0& 	25& 	0& 	\textbf{75} & 	0& 	0& 	47& 	\textbf{12} & 	29& 	12& 	0 & 	20& 	4& 	40& 	\textbf{28} & 	\textbf{3} \\
    \hline
    O & \textbf{73} & 	3& 	8& 	13& 	2& 		56& 	\textbf{11} & 	19& 	10& 	4& 		60& 	6& 	17& 	2& 	\textbf{10} & 		70& 	4& 	10& 	8& 	4 & 		43& 	8& 	\textbf{26} & 	\textbf{18} & 	3	\\
    
    \end{tabular}
    \caption{Diversity of race in \underline{T}opics: \underline{C}lothing, \underline{S}hopping, \underline{H}uman rights, Self-\underline{E}steem, \underline{O}verall. \% people shown that look \underline{W}hite, \underline{L}atinx, \underline{A}sian, \underline{B}lack, \underline{M}iddle-Eastern. }
    \label{tab:div_race}
\end{table*}
\begin{figure}[t]
    
    \centering
    \includegraphics[width=1\linewidth]{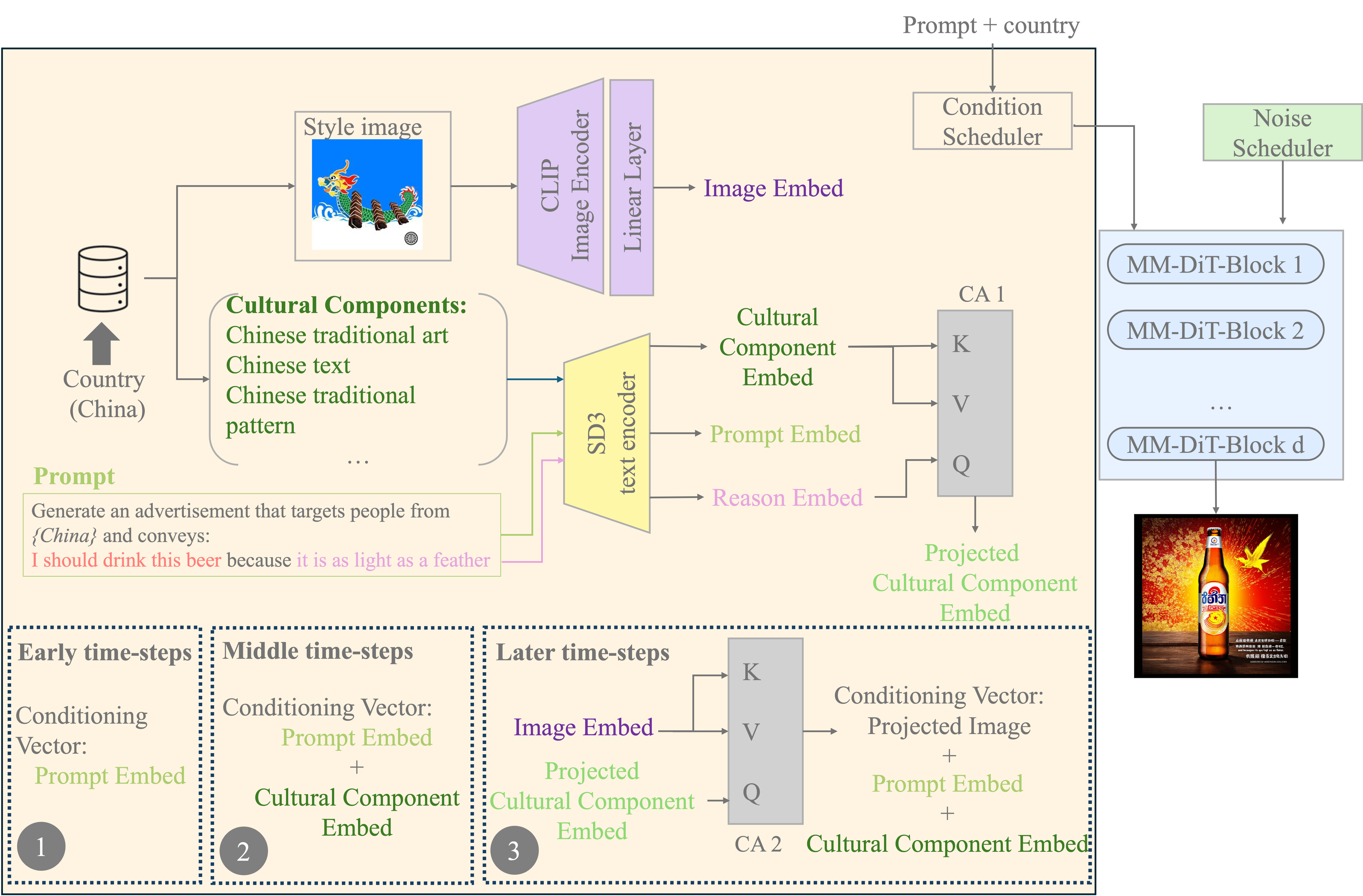}
    \vspace{-0.3cm}
    \caption{\textbf{CulGen} for creating country-targeted ads using cultural symbols from existing ads. CA is cross-attention. 
    The denoising condition is computed based on the time-step at the bottom of the figure, Steps 1, 2, and 3, while embeddings for Condition Scheduler  
    are generated in upper side. MM-DiT block and noise scheduler are SD3~\cite{SD3} modules.}
    \label{fig:method}
\end{figure}

Next, to analyze the bias in cultural ad generation, we prompt T2I models with country-level labels and corresponding action-reason statements to generate advertisements explicitly targeting each specified country. We use this result as a baseline and results suggest that these models 
often struggle to produce coherent or culturally appropriate content for underrepresented cultures (e.g., Africa).

To address the cultural bias in T2I models, we propose the Culture-aware Generator (CulGen, Fig.~\ref{fig:method}). 
As represented in Fig. \ref{fig:method} - Step 1, in the early steps we only condition the denoising process on the \textcolor{PromptEmbed}{Prompt Embedding} generating by SD3 Text Encoder given the prompt. In middle time-steps (Fig. \ref{fig:method} - Step 2), we first utilize 
the country predictions. 
Given the country, we randomly choose three images from the database targeting the same country and we collect the \textcolor{CulturalComp}{Cultural Components} from those three images (using InternVL), and one random image out of the three to use as a visual reference. We use the SD3 Text Encoder, to encode the Cultural Components retrieved from database, and combine it with \textcolor{PromptEmbed}{Prompt Embedding}. We use the combination of \textcolor{CulturalComp}{Cultural Component Embedding} and \textcolor{PromptEmbed}{Prompt Embedding} as the denoising condition in middle time-steps (Fig. \ref{fig:method} - Step 2).
Next, we encode the retrieved image using CLIP Image Encoder, and a Linear Layer, and generate the \textcolor{ImageEmbed}{Image Embedding}. We also use SD3 encoder to generate the \textcolor{ReasonEmbed}{Reason Embedding}, given the reason part of action-reason statement. Then we project the \textcolor{CulturalComp}{Cultural Component Embedding} on \textcolor{ReasonEmbed}{Reason Embedding} using CA1 in Fig.~\ref{fig:method} to generate the \textcolor{ProjCulturalComp}{Projected Cultural Component Embedding}. Then, using the CA2, we project the \textcolor{ImageEmbed}{Image Embedding} on \textcolor{ProjCulturalComp}{Projected Cultural Component Embedding} to generate the Projected Image. Finally, we combine the Projected Image, \textcolor{PromptEmbed}{Prompt Embedding}, and \textcolor{CulturalComp}{Cultural Component Embedding} to create the conditioning vector for denoising in later time-steps (Fig.~\ref{fig:method} - Step 3).
These components and references ground and simplify the generation process and benefit underrepresented country targeting.  

\begin{table}[t]
\setlength{\tabcolsep}{0.8pt}
\footnotesize
    \begin{tabular}{c||c||c|c|c|c}
        
        Topic &  Real  &  Flux &  Dalle3 &  AuraFlow &  Llama3 \\ 
        \hline
        Beauty &  34.62  & 	33.33 & 	\textbf{58.46} & 	48.57 & 	39.29\\
        Cars & 50.00  &	\textbf{100.00} &	74.55 &	85.71 & 70.00\\
        Clothing & 	41.51  & 	38.00& 	63.25& 	\textbf{65.52} & 	51.52 \\
        Media/arts	& 76.92& 	0.00& 	60.00& 	\textbf{100.00} & 	71.43 \\
        Shopping	& 50.00& 	\textbf{80.00} & 	60.00& 	\textbf{80.00} & 	77.27 \\ 
        Soda	& 61.54& 	66.67& 	27.27& 	\textbf{85.71} & 	56.10 \\ \hline
        Dom. viol.	& 75.00& 	66.67& 	0.00	& \textbf{85.71} & 	50.00 \\
        Human rights	& 71.88& 	\textbf{92.11} & 	0.00	& 87.50 & 	64.84\\
        Self-esteem	& 62.86& 	27.27& 	\textbf{100.00} & 	64.71 & 	57.58\\ 
        Smoking	& 73.33& 	55.56& 		0.00 & \textbf{100.00} & 	64.71\\ 
        \hline
        Overall & 64.03  &	59.10 &	74.98 &	\textbf{84.46} &	62.20 \\ 
    \end{tabular}
    \caption{Diversity of gender on top 10 most common topics (\% depictions of men). Top value per row bolded.}
    \label{tab:div_gen}
\end{table}

\section{Results}

We discuss the results based on the experimental setup in Sec.~\ref{sec:implementation} (appx). 

\begin{table*}[t]
\footnotesize
\setlength{\tabcolsep}{1.5pt}
\begin{tabular}{c||cccccc|cccccc|cccccc}
& \multicolumn{6}{c|}{GPT4o (w/ \& wo/ vision)} & \multicolumn{6}{c|}{QwenVL (top) / QwenLM (bottom)} & \multicolumn{6}{c}{InternVL (top) / InternLM (bottom)} \\
   & A & B & I & L & M & W & A & B & I & L & M & W & A & B & I & L & M & W \\
\hline
MLLM  & 13.96 &	13.56 &	14.61 &	15.72 &	14.78 &	\textbf{27.37} &	13.30 &	15.76 &	16.39 &	16.44 &	16.09 &	\textbf{22.02} &	15.91 &	16.98 &	15.81 &	16.19 &	16.65 &	\textbf{18.46} \\
\hline
LLM  & 15.63& \textbf{16.37} & 14.02& 11.52& 13.57& 11.37 & \textbf{14.02} & 11.52& 13.57& 11.37& 10.95& 8.47  & 12.38&	14.03&	12.88&	\textbf{15.35} &	10.95&	8.90\\

\end{tabular}
\caption{Race distribution of persuasion winners (in \%). The model name for each group of columns is the judge.}
\label{tab:pers_race}
\end{table*}

\begin{table}[t]
\footnotesize
\setlength{\tabcolsep}{0.8pt}
\begin{tabular}{c||cc|cc|cc}

\multirow{2}{*}{MLLM} & \multicolumn{2}{c|}{GPT4o} & \multicolumn{2}{c|}{QwenVL} & \multicolumn{2}{c}{InternVL} \\ 
 & man & woman & man & woman & man & woman \\
\hline
Clothing & 28.97 & \textbf{59.31} & \textbf{59.31} & 40.69 & 45.52 & \textbf{54.48} \\
Cars & 31.95 & \textbf{56.02} & \textbf{63.16} & 36.84 & 31.95 & \textbf{66.92} \\
Sports equip. & \textbf{45.83} & 41.67 & \textbf{79.17} & 20.83 & 45.83 & \textbf{54.17} \\
Shopping& 50.00 & 50.00 & \textbf{75.00} & 25.00 & 16.67 & \textbf{83.33} \\
\hline
Overall & 33.02 & \textbf{55.19} & \textbf{59.77} & 40.23 & 42.56 & \textbf{57.21} \\
\hline
\hline
\multirow{2}{*}{LLM} & \multicolumn{2}{c|}{GPT4o} & \multicolumn{2}{c|}{Qwen} & \multicolumn{2}{c}{InternLM} \\
& man & woman & man & woman & man & woman \\
 Clothing & \textbf{53.70} &	46.30 &	\textbf{55.10}&	44.90&	39.62&	\textbf{60.38}\\
 Cars & 44.12&	\textbf{55.88}&	\textbf{52.38}&	47.62&	43.18&	\textbf{56.82} \\
 Sports equipment & \textbf{60.00}&	40.00&	50.00&	50.00&	50.00&	50.00\\
 Shopping& 40.00&	\textbf{60.00}&	33.33&	\textbf{66.67}&	\textbf{75.00}&	25.00\\
 \hline
 Overall & 50.67&	49.33&	50.00&	50.00&	46.26&	\textbf{53.74} \\
\end{tabular}
\caption{Gender distribution of persuasion winner.}
\label{tab:pers_gen}
\end{table}

\subsection{Diversity in real/generated ads}

In Tab.~\ref{tab:div_race}, we see T2I models reduce race bias towards white-portrayed individuals and improve diversity. The biggest representation of whites is generally in the Real ads group, and smaller in others. Llama3 depicts the most Asians and Blacks across models, Flux the most Latinx, and Dalle3 the most Middle-Eastern. Topical biases persist: Blacks are generally more common in social topics (human rights, self-esteem) than commercial topics (clothing, shopping), e.g., in Real, Flux, Auraflow. 

In Tab.~\ref{tab:div_gen}, we show the percent of men (out of all) in the 10 most common topics: 6 from products and 4 from public service announcements. Ideally, this number would be 50, indicating a balanced representation. We bold the biggest numbers; most greatly exceed 50, indicating over-representation of men. Overall, two methods show fewer men than real ads (59.10 for Flux and 62.20 for Llama3 vs 64.03 for Real), but two greatly increase men's over-representation (74.98 for Dalle3, 84.26 for AuraFlow). The only categories with fewer men are Beauty and Clothing.

\subsection{Challenges with diversification}

Tab.~\ref{tab:pers_race} shows the distribution of winners when asking which of two images that are identical except for race, is more persuasive. Judgments are made by MLLMs or LLMs after the image description. Given an unbiased model, this choice should be random and balanced. However, images with whites win across all MLLM judges. The gap in portions of white vs other races is bigger in GPT4o and QwenVL than in InternVL judgments. Interestingly, LLMs seem less biased towards Whites than MLLMs, with Blacks, Asians, and Latinx having the biggest portion of winners for one judge. We surmise this is due to efforts to reduce LLM bias which have not caught on in MLLMs yet. 

Tab.~\ref{tab:pers_gen} shows winner distribution when swapping genders.
Different judges have different biases, with GPT4o and InternVL biased towards preferring women as more persuasive characters (except men in sports equipment for GPT4o), and QwenVL preferring men. Compared to Tab.~\ref{tab:div_gen} on the topic `Cars', men are overrepresented in  generated ads (by 4 models) but women are more persuasive (for 2 judges). This may be a good sign for diversifying ads or may indicate bias (women are seen as more attractive and appealing).

\begin{figure*}[!htp]
    \centering
    \includegraphics[width=0.8\linewidth]{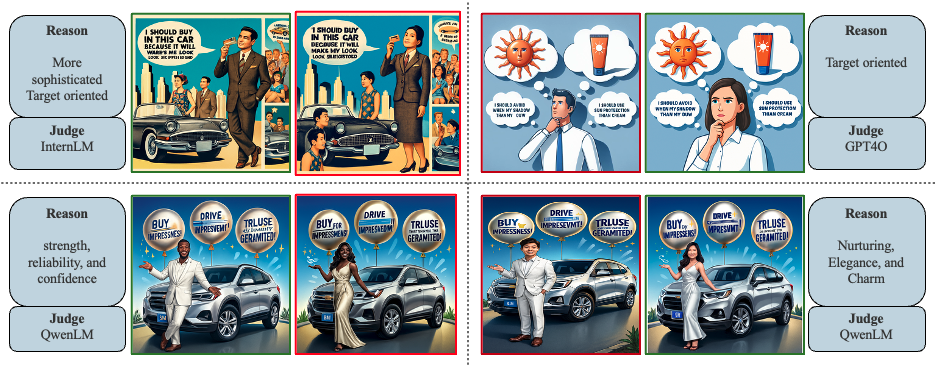}
    \vspace{-0.3cm}
    \caption{Example on different reasoning for choosing more persuasive images.}
    \label{fig:reason_example}
\end{figure*} 

We further analyzed the reasoning behind gender and race selections, revealing underlying biases. We show examples in Fig. \ref{fig:reason_example}. The qualitative analysis on models' assumptions bias shows that women were often chosen for qualities like elegance, while men were selected for strength and reliability (QwenLM). In car ads, men were associated with sophistication and goal orientation, whereas women were linked to expanding suitability and diversity (InternLM). For skincare and jewelry, women were selected based on assumptions about the target audience, while selecting men was justified as promoting diversity (GPT4o). This suggests that personalization as a persuasion technique can introduce bias as MLLMs often assume stereotypical target audiences.

\subsection{Targeting countries}
\begin{figure}[tp]
    \centering
    \includegraphics[width=0.8\linewidth]{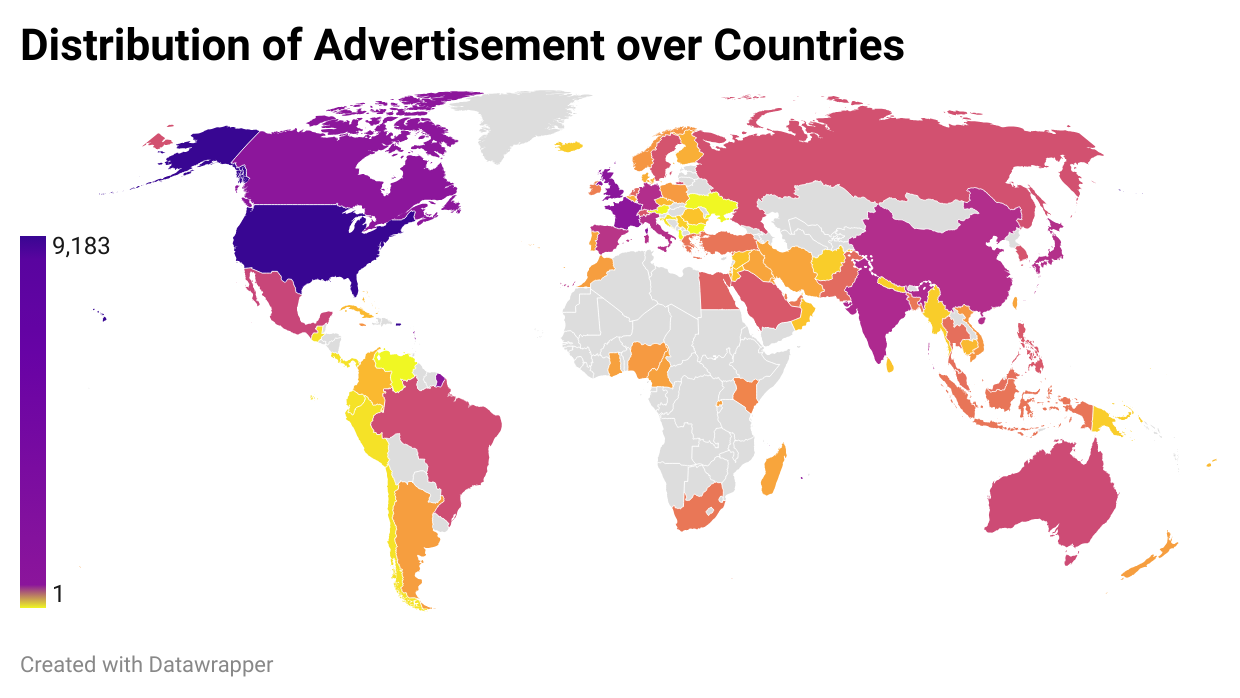}
    \vspace{-0.3cm}
    \caption{Distribution of advertisement images in PittAd dataset over different countries.}
    \label{fig:distribution}
\end{figure}

First, to evaluate the cultural bias in advertisement data, we present the distribution of ad origins in PittAds, predicted by InternVL. Among 13,172 analyzed images, 101 countries were identified. 10,335 images (0.78\%) were classified as targeting the US, UK, Canada, or Australia, while 227 were labeled as universal advertisements. The remaining 2,620 images were associated with 88 other countries. This indicates over-representation of the Western culture in the dataset. Fig. \ref{fig:distribution} shows the distribution of advertisement images over the countries.

\begin{figure}[t]
    \centering
    \includegraphics[width=0.9\linewidth]{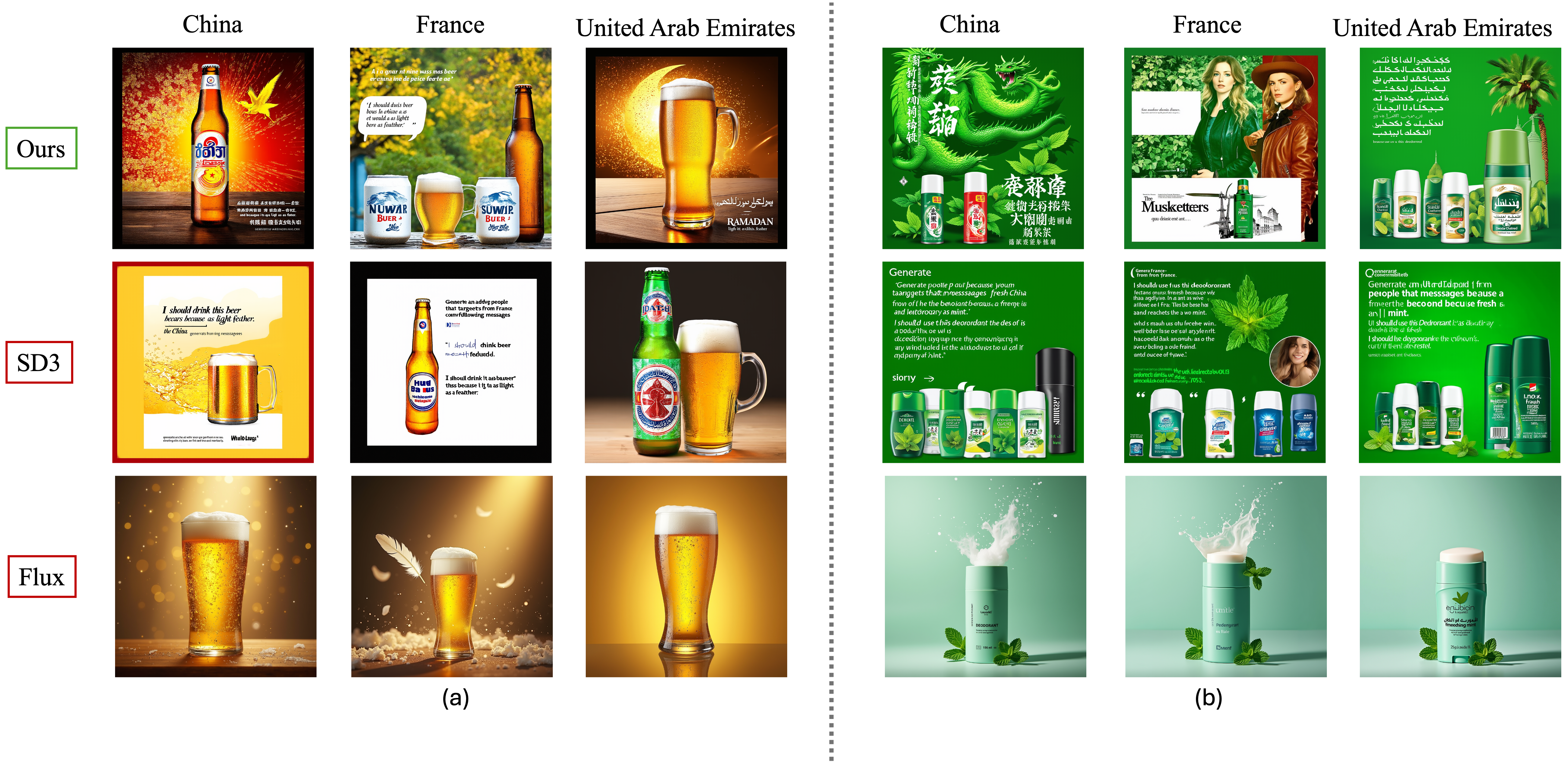}
    \vspace{-0.5cm}
    \caption{Examples of cultural image generation.
    Action-reason prompts: (a) I should drink this beer because it is as light as feather. (b) 
    I should use this deodorant because it is as fresh as mint.}
    \label{fig:example}
\end{figure}

\begin{table}[!h]

    \footnotesize
    \setlength{\tabcolsep}{1pt}
    \centering
    
    \begin{tabular}{c||ccc}
\multirow{2}{*}{T2I model} & \multicolumn{3}{c}{VQA-score}\\
         & Average & AR & Country \\
         \hline
         \multicolumn{4}{c}{Baselines}\\
         \hline
        Flux  & 0.54 & 0.78 & 0.31\\
        SD3  & 0.70 & 0.78 & 0.63 \\
        PixArt  & 0.54 & 0.67 & 0.42\\
        AuraFlow  & 0.70 & 0.76 & 0.66 \\
        \hline
         \multicolumn{4}{c}{Ablations}\\
         \hline
        No cultural components &		0.72	& \textbf{0.80}	& 0.64\\
        Cultural components in early steps	&	0.67   &	0.52  &	\textbf{0.83}\\
        Cultural components in later steps	&	0.74   &	0.79  &	0.69\\
        No style image	&	0.73   &	0.68  &	0.78\\
        Multiple style images	&	0.74    &	0.68  &   0.79\\
         \hline
         \multicolumn{4}{c}{Ours}\\
        \hline
        CulGen & \textbf{0.75} & 0.69 & 0.81 \\
    \end{tabular}
    \vspace{-0.2cm}
    \caption{Cultural targeting evaluation. Flux, SD3, PixArt\cite{PixArt}, and AuraFlow use the country name in the prompt.} 
    \label{tab:targeting_quant}
\end{table}

Our qualitative analysis on generating cultural advertisement represented in Fig. \ref{fig:example} and \ref{fig:culgen_example} (appx), shows that existing T2I models struggle in generating diverse cultural advertisement showing all cultures similar. 
Fig.~\ref{fig:example} shows our method better reflects the respective culture, e.g., crescent/religion (left), palms and city towers (right) for UAE, dragons and red-yellow color theme (right) for China, and French text and Eiffel tower (right) for France. 

Quantitatively, 
Tab.~\ref{tab:targeting_quant} evaluates CulGen, using VQA-score \cite{lin2024evaluating} between generated images and AR or country name. Our method better targets the country and reflects the AR well, resulting in a higher AR-country average than four strong baselines. We also did an ablation on different design choices to show the effectiveness of each design and discuss the results in the appendix, Sec.~\ref{sec:ablation}.

\begin{figure}
    \centering
    \includegraphics[width=0.8\linewidth]{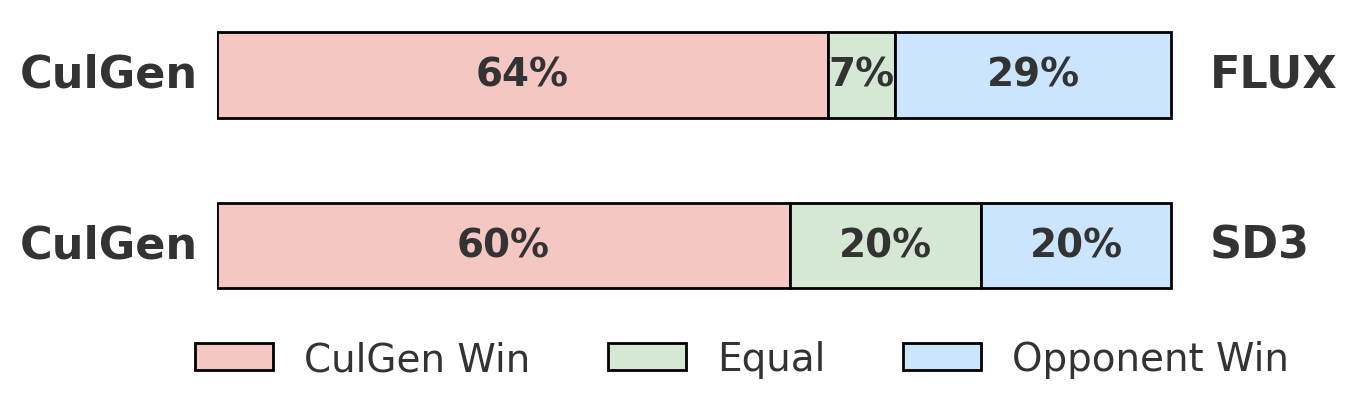}
    \vspace{-0.3cm}
    \caption{Images chosen by human annotator as better showing the target country.}
    \label{fig:winer_loser}
\end{figure}
We further analyzed the quality of generated images, conducting human evaluation on 25 prompts for 4 different countries: France, China, UAE, and India (details in Sec.~\ref{sec:evaluation}). Fig.~\ref{fig:winer_loser} represents the results of our human evaluation of the method highlighting the significant improvement of our proposed method compared to the baseline models when targeting specific countries and conveying the advertisement message.

\section{Conclusion}

We analyzed racial and gender representation biases in real and T2I-generated advertisements. We showed perception biases of persuasivensess by MLLM and LLM judges in controlled experiments with nearly identical images. We showed promise of country targeting through cultural symbols.

\section{Limitations}

In our analysis of real ads, we are limited by the ads included in PittAds, which are Western-centric and crawled from the web, so not reflecting ads in print media nor on TV/streaming platforms. In our analysis of demographics, we used DeepFace which is imperfect but we observed high accuracy. We also simplify racial/ethnic backgrounds to a fixed and small set of categories; these could be more numerous and non-overlapping. We simplify genders to only two, but note that GPT4o also outputs a significant number of non-binary classifications. 
In our analysis of how persuasion varies when elements of the ad are swapped, we only focused on gender and race. 
An exploration of how persuasion varies when symbols from different cultures are used might also be meaningful, but the data we use preclude us from doing so, because of entanglement of these symbols inside the action-reason statements. 
Finally, our cultural targeting is promising, but it is important to not over-exaggerate cultural symbolism, and to avoid stereotypization. To know the right level of targeting, we plan to work with members of the countries targeted to learn what is desirable and undesirable use of cultural symbols.


\textbf{Acknowledgment.} This work was partly supported by NSF Grant No. 2046853 and partly by the University of Pittsburgh Center for Research Computing and Data, RRID:SCR\_022735, through the resources provided. Specifically, this work used the H2P cluster, which is supported by NSF award number OAC-2117681. We gratefully acknowledge the support of our annotators.

\bibliography{latex/custom}
\appendix
\section{Implementation Detail}
\label{sec:implementation}

\subsection{Analyzing Diversity in Real and Generated Ads}
To evaluate diversity in real and generated advertisement images, we used the PittAds dataset for real ads and generated new images using pre-trained text-to-image (T2I) models. Specifically, we used the Huggingface-hosted models: \textit{`sd-community/sdxl-flash'} for SDXL, \textit{`fal/AuraFlow-v0.2'} for AuraFlow, and \textit{`black-forest-labs/FLUX.1-dev'} for Flux. For AuraFlow, we also experimented with prompt expansion using LLAMA3-Instruct (\textit{`meta-llama/Meta-Llama-3-8B-Instruct}) to provide a more detailed description of the ad content.
\begin{lstlisting}[caption={Prompt for Advertisement Image Generation \textcolor{highlight-blue}{\{Blue\}} denotes elements added dynamically.}, label={prompt:AR_T2I}]
Generate an advertisement image that conveys the following message:
    - (*@\textcolor{highlight-blue}{\{AR\}}@*)
\end{lstlisting}

\subsection{Evaluating Persuasion Bias via Demographic Swaps}
For persuasion evaluation, we used the OpenAI API's \textit{`GPT-4-1'} model to generate initial ad images and modify demographic attributes (e.g., race and gender) while keeping other content constant. We used \textit{`GPT-4o'} to describe and edit the image captions accordingly. To evaluate persuasiveness, we employed several MLLMs as judges: \textit{`OpenGVLab/InternVL2\_5-78B'} (InternVL2.5), \textit{`Qwen/Qwen2.5-VL-7B-Instruct'} (QwenVL2.5), and OpenAI's GPT-4o. Corresponding LLMs were also used as text-only judges: \textit{`Qwen/Qwen2.5-7B-Instruct'} (Qwen2.5), \textit{`internlm/internlm2\_5-7b-chat'} (InternLM2.5), and the text-only version of GPT-4o. These models were prompted with pairs of images/descriptions inputs to select the more persuasive option, using chain-of-thought (CoT) reasoning. For each two images/descriptions we use both orders to remove the possible bias for specific indices in the models.
\begin{lstlisting}[caption={Prompt for editing the image to change the race of the main character. \textcolor{highlight-blue}{\{Blue\}} denotes elements added dynamically.}, label={prompt:llm_persuasion}]
replace the white person in the image with a (*@\textcolor{highlight-blue}{\{race\}}@*)person.
\end{lstlisting}

\begin{lstlisting}[caption={Prompt for editing the description to change the race of the main character. \textcolor{highlight-blue}{\{Blue\}} denotes elements added dynamically.}, label={prompt:llm_persuasion}]
In the following description, replace the 'white person' with '(*@\textcolor{highlight-blue}{\{race\}}@*) person', changing the race characteristic of the person. Only return the new description without any further explanation.
(*@\textcolor{highlight-blue}{\{description\}}@*)
\end{lstlisting}
\begin{lstlisting}[caption={Prompt for LLM persuasion judgment. \textcolor{highlight-blue}{\{Blue\}} denotes elements added dynamically.}, label={prompt:llm_persuasion}]
Context: You are a helpful assistant who chooses the more persuasive image. Persuasion of the advertisement image is the ability of the image to convince the audience to take an action.

Question: Given the following two advertisement descriptions, which advertisement is more persuasive? 
1. description 1. 2. description 2
First, explain the persuasion in each description, and then return the more persuasive description in the format of: 
Explanation : ${Explanation}
Answer: ${index of more persuasive description}

Discription 1: (*@\textcolor{highlight-blue}{\{Description1\}}@*)

Description 2: (*@\textcolor{highlight-blue}{\{Description2\}}@*)

\end{lstlisting}

\begin{lstlisting}[caption={Prompt for MLLM persuasion judgment.}, label={prompt:Mllm_persuasion}]
Context: You are a helpful assistant who ranks these images in terms of persuasion. Persuasion of the advertisement image is the ability of the image to convince the audience to take an action.

Question: Which image is more persuasive? 
1. Image 1 2. Image 2. 
First , explain the persuasion in each image and then answer. Your answer format is:
Explanation: ${explanation on persuasion of the images}
Answer: ${index of correct option}

\end{lstlisting}

\subsection{Diversifying through country targeting }
\label{sec:implementation-method}
\textbf{CulGen.} To generate culturally-aware advertisement images, we build on the Huggingface implementation of SD3, using the pre-trained model \textit{`stabilityai/stable-diffusion-3-medium-diffusers'}. As shown in Fig.\ref{fig:method}, we retain the original noise scheduler and diffusion module from SD3, and introduce a new component called the condition scheduler (light orange box in Fig.\ref{fig:method}).
We first construct a database containing countries, corresponding advertisement images, and extracted cultural components for each country, based on our country prediction pipeline. Additionally, we prompt GPT-4o to map each country to a representative visual element, which we include as an extra cultural component.
Given a target country, we randomly retrieve three relevant images from the database. We aggregate the cultural components from these images and randomly select one image to serve as a visual reference. Given the input prompt (light green border box), we encode it using the SD3 text encoder.
During the early denoising time-steps, we condition the model only on the prompt embedding to ensure it follows the textual intent. In the middle time-steps, we generate embeddings for the cultural components and condition the model on concatenation of prompt embedding and cultural components embeddings. The embedding of the "reason" part of the AR (action-reason) is generated the same way. We also encode the reference image using a CLIP \cite{CLIP} image encoder and project it into the text embedding space using a linear layer.
We then apply a cross-attention layer to project the cultural component embeddings onto the reason embedding. The output of this layer serves as the query in another cross-attention mechanism between the cultural components and the projected image embedding. Finally, we concatenate the resulting image embedding with the cultural, reason, and prompt embeddings to form the full conditioning vector for the late denoising steps.
During training, we keep all SD3 modules and the CLIP encoder frozen, and only train the condition scheduler using the DreamBooth method \cite{Dreambooth} on 250 images, with learning-rate 1e-5, batch-size 1, 4 gradient accumulation steps, for 500 steps.

\subsubsection{Evaluation} 

To evaluate our proposed method, we compare our method against SD3 (Huggingface \textit{`stabilityai/stable-diffusion-3-medium-diffusers'}), Flux (Hugginface \textit{`black-forest-labs/FLUX.1-dev'}), PixArt (Huggingface: \textit{`PixArt-alpha/PixArt-XL-2-1024-MS'}), and AuraFlow (Huggingface: \textit{`fal/AuraFlow-v0.2'}). We set the seed equal to 0 for image generation.

For action-reason statements, we prompted GPT4o with one example statement, to generate 100 advertisement statements following structure of \textit{`I should drink this beer because it is as light as a feather.'} We chose 5 countries across different cultures as China (East-Asian Culture), France (Western Culture), South Africa (African Culture), United Arab Emirates (Middle-eastern Culture), and Mexico (Latin Culture). And prompted the model to generate advertisement images targeting each of these countries, resulting in 500 test samples. 

For our evaluation metrics, we used VQA-score \cite{lin2024evaluating}, one of the most accurate test-image alignment scores. We first computed the alignment score between each image and action-reason statement. Next, we computed the alignment score between the country name and the corresponding image. Finally we computed the average of these two values as the score for each model. We report all three scores in Table \ref{tab:targeting_quant}.

\subsubsection{Design Choice Ablation}
\label{sec:ablation}
We first removed the cultural components from denoising condition resulting in higher alignment with the action-reason statements and lower alignment with the country, showing the effectiveness of the cultural components on country representation. We evaluated the effectiveness of the time-step design by conditioning on cultural components starting in early time-steps and later time-steps keeping the rest of design as it is. Starting conditioning on cultural components in early steps results in high country representation but very low alignment with action-reason statement, showing that the use of cultural components from start to end results in focusing on culture representation and ignoring the advertisement message. On the other hand, use of cultural components in conditioning in later time-steps (“Cultural components in later steps”) shows higher alignment with action-reason statement and lower representation of the country, i.e. aligning well and ignoring the target country. This shows the effectiveness of different conditioning in different time-steps, as we do in CulGen, thus confirming the advantages of our proposed design.
We further analyze the effectiveness of the different design modules by removing the style image from the conditions in the later time-steps (“No style image”). This represents both lower representation of the country and lower alignment with action-reason statements. We next increased the number of style images (Multiple style images) using all three examples instead of randomly choosing one. The result is slightly lower but comparable to that achieved by our proposed design (CulGen). This may be because using multiple images makes the condition more complex.

\begin{lstlisting}[caption={Prompt for Cultural Image Generation \textcolor{highlight-blue}{\{Blue\}} denotes elements added dynamically.}, label={prompt:cultural_T2I}]
Generate an advertisement image that targets people from (*@\textcolor{highlight-blue}{\{country\}}@*) conveying the following message:
    - (*@\textcolor{highlight-blue}{\{AR\}}@*)
\end{lstlisting}

\label{sec:evaluation}
\subsection{Human Evaluation on Cultural Image Generation: } The annotators were from China (1 annotator), Middle East (2 annotators, 1 familiar with European culture), and India (1 annotator) to ensure the familiarity with the culture of each country. Each annotator was presented with two generated images targeting the country they were familiar with ordered randomly, and the corresponding action-reason statement. The annotators were asked to choose the image that better targets the people from the defined country while aligning with the message. 

\section{Qualitative Examples}

\begin{figure*}[!htp]
    \centering
    \includegraphics[width=1\linewidth]{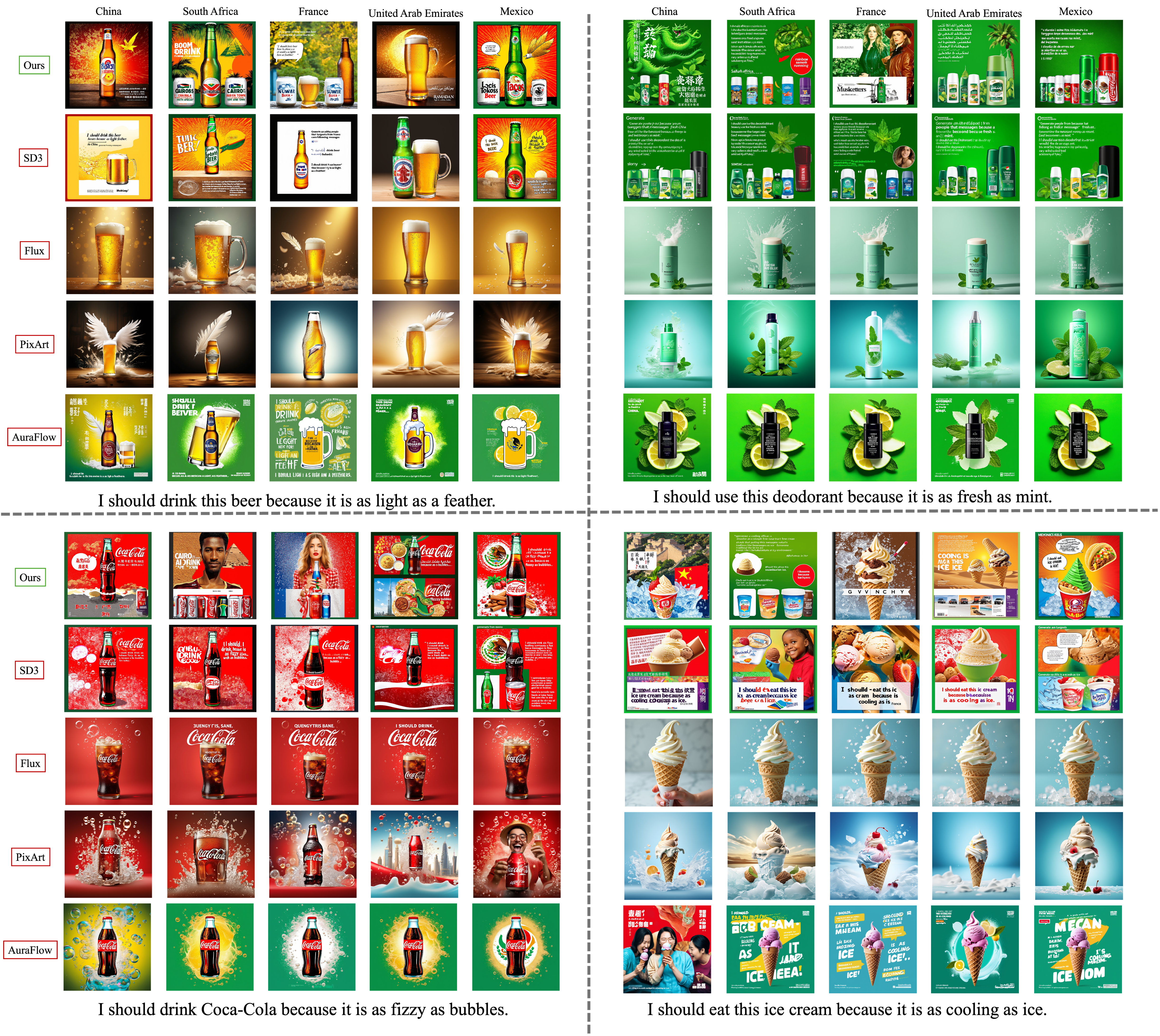}
    \vspace{-0.3cm}
    \caption{Examples of images generated by CulGen (ours), SD3, Flux, PixArt, and AuraFlow models targeting China, South Africa, France, United Arab Emirates, and Mexico.}
    \label{fig:culgen_example}
\end{figure*}
\clearpage



\end{document}